\documentclass{article}

\usepackage{arxiv}
\usepackage[utf8]{inputenc} % allow utf-8 input
\usepackage[T1]{fontenc}    % use 8-bit T1 fonts
\usepackage{hyperref}       % hyperlinks
\usepackage{url}            % simple URL typesetting
\usepackage{booktabs}       % professional-quality tables
\usepackage{amsfonts}       % blackboard math symbols
\usepackage{nicefrac}       % compact symbols for 1/2, etc.
\usepackage{microtype}      % microtypography
\usepackage{lipsum}
\usepackage{graphicx}
\usepackage{notoccite}
\usepackage{subcaption}
\usepackage{array}
\usepackage{multirow} % TO MERGE ROWS IN A TABLE
\usepackage{graphicx}
\usepackage{adjustbox}
\usepackage{wrapfig}
\usepackage{mathpazo} % Palatino font
\usepackage[hang,flushmargin]{footmisc}
\usepackage[style=ieee]{biblatex} 
\usepackage{amsmath}
\usepackage{makecell}
\usepackage{units}
\usepackage{appendix}
\usepackage{pdflscape}

\hypersetup{hidelinks} % Hide colored boxes in crossrefs

\graphicspath{ {./Figures/} }

\title{Artificial Intelligence in Brazilian News: A Mixed-Methods Analysis}

\date{\today}

\author{ \href{https://orcid.org/0009-0005-0323-8326}{\includegraphics[scale=0.06]{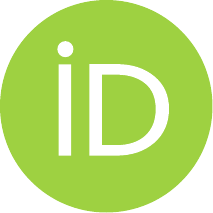}\hspace{1mm}Raphael Hernandes} \\
	Leverhulme Centre for the Future of Intelligence\\University of Cambridge\\
	\texttt{rhh43@cam.ac.uk}
    \And % This is the separator for the next author
    \href{https://orcid.org/0000-0002-5130-2258}{\includegraphics[scale=0.06]{orcid.pdf}\hspace{1mm}Giulio Corsi}\\ % Replace with the second author's name
    Leverhulme Centre for the Future of Intelligence\\University of Cambridge\\
    \texttt{gc540@cam.ac.uk} % Replace with the second author's email
}

\renewcommand{\headeright}{}
\renewcommand{\undertitle}{}
\renewcommand{\shorttitle}{}

\hypersetup{
pdftitle={Artificial Intelligence in Brazilian News: A Mixed-Methods Analysis},
pdfsubject={ethics of ai},
pdfauthor={Raphael Hernandes, Giulio Corsi},
pdfkeywords={news coverage, artificial intelligence, Brazilian media, computational grounded theory, topic modeling, media bias},
}

\begin{document}
\maketitle

\begin{abstract}

    The current surge in Artificial Intelligence (AI) interest, reflected in heightened media coverage since 2009, has sparked significant debate on AI's implications for privacy, social justice, workers' rights, and democracy. The media plays a crucial role in shaping public perception and acceptance of AI technologies. However, research into how AI appears in media has primarily focused on anglophone contexts, leaving a gap in understanding how AI is represented globally. This study addresses this gap by analyzing 3,560 news articles from Brazilian media published between July 1, 2023, and February 29, 2024, from 13 popular online news outlets. Using Computational Grounded Theory (CGT), the study applies Latent Dirichlet Allocation (LDA), BERTopic, and Named-Entity Recognition to investigate the main topics in AI coverage and the entities represented. The findings reveal that Brazilian news coverage of AI is dominated by topics related to applications in the workplace and product launches, with limited space for societal concerns, which mostly focus on deepfakes and electoral integrity. The analysis also highlights a significant presence of industry-related entities, indicating a strong influence of corporate agendas in the country's news. This study underscores the need for a more critical and nuanced discussion of AI's societal impacts in Brazilian media.

\end{abstract}

\section{Introduction}

Artificial Intelligence (AI) hype is at an unprecedented level~\parencites{markeliusMechanismsAIHype2024}. This keen interest reflects and is reflected in the news. In \textit{The New York Times}, for example, there has been increased AI coverage since 2009~\parencites[964]{fastLongTermTrendsPublic2017}. Journalism reacts to readers' attention to a topic but also sets the agenda of public debate~\parencites{weaverMediaAgendaWho2014,Tandoc2021,macgregorTrackingOnlineAudience2007}.

AI comes with concerns about its impacts on privacy, social justice, workers' rights, and democracy. The discourse about AI in media, fiction or non-fiction, helps set public perception, acceptance, expectations, and how the technology will be implemented~\parencites[5--6]{caveIntroductionImaginingAI2020}[]{brennenWhatExpectWhen2022}[]{dillonWhatAIResearchers2023}. Given its importance, researchers have analyzed how AI appears in different mediums and the effects these portrayals have~\parencites{caveIntroductionImaginingAI2020,caveAIRaceStrategic2018,caveHowWorldSees2023,recchiaFallRiseAI2020,brennenIndustryledDebateHow2018}. 

News coverage of AI, however, remains underexplored~\parencites{brennenIndustryledDebateHow2018}, as are studies exploring AI representation beyond the anglophone world~\parencites[5]{caveHowWorldSees2023}. Looking into different locations is crucial, as the attitudes towards AI and its coverage vary around the globe~\parencites[4]{caveHowWorldSees2023}[]{wangAIAnxietyComparing2023}. Hence, this study addresses both gaps by bringing new empirical evidence about how AI is covered in Brazilian news. 

Brazil is the largest Portuguese-speaking country and is situated in the Global South, providing an interesting focus of study beyond the anglophone tradition. This is supported by its unique history of rapid technological adoption in the 2000s~\parencites[169--170]{kingAfrofuturismoAestheticsResistance2023}, and digital platforms' strong influence in the country's democracy~\parencites{melloMaquinaOdioNotas2020,ozawaHowDisinformationWhatsApp2023}.

This study will analyze 3,560 news articles published between July 1\textsuperscript{st}, 2023, and February 29\textsuperscript{th}, 2024. They come from 13 of the country's most popular online news outlets, which can be subdivided into 21 publications (due to cases when the same company owns more than one outlet) of different characteristics, including legacy media (of national and state relevance), large news portals, and websites specialized in economy and in technology. This research aims to answer the following questions:

\begin{itemize}
    \item[RQ1] What are the main topics in Brazilian news coverage of AI?
    \item[RQ2] Who are the entities (people, companies) represented in the news about AI in Brazil?
\end{itemize}

These questions will help understand Brazilian news' attitude towards AI. They will be explored through Computational Grounded Theory (CGT), a framework that combines computational and qualitative methods to rigorously analyze content~\parencites{nelsonComputationalGroundedTheory2020}. The implementation for RQ1 will include a progression of steps to identify topics, going through machine learning techniques, with Latent Dirichlet Allocation (LDA)\footnote{A probabilistic, unsupervised machine learning technique used to discover abstract topics within a collection of documents by analyzing the distribution of words in the texts~\parencites[48]{radimrehurekSoftwareFrameworkTopic2010}.} and BERTopic\footnote{An advanced modeling technique using BERT embeddings and clustering algorithms to identify topics in texts~\parencites{grootendorstBERTopicNeuralTopic2022}.}, and qualitative analysis. RQ2 will be explored with Named-Entity Recognition (NER)\footnote{Technique that identifies entities in texts and classifies them as a person, organization, location, or miscellaneous~\parencites{melaniewalshNamedEntityRecognition2021}.}. These techniques have already been proven valuable to extract meaning from large corpora of news articles~\parencites{varolUnderstandingCOVID19News2022,chipidzaTopicAnalysisTraditional2022,guntherWordCountsTopic2018b,hamborgAutomatedIdentificationMedia2019}, content related to AI~\parencites{recchiaFallRiseAI2020,weberHowMedicalProfessionals2024,fastLongTermTrendsPublic2017}, and papers about AI in journalism in Brazil~\parencites{ioscoteJornalismoInteligenciaArtificial2021}.

This is the first work to explore how AI is covered in Brazilian news. Literature reviews show that previous studies on the intersection between AI and journalism in the country focused on how newsrooms can apply AI and its impacts on the journalists' \textit{praxis}~\parencites[175--178]{ioscoteJornalismoInteligenciaArtificial2021}[13]{moisescostapintoInteligenciasArtificiaisNos2021}.

Portuguese terms will be translated into English by the authors. Translated terms will be written in \textsc{small caps}. A list of the original terms and translations can be found in Appendix~\ref{sec:terms}.

\section{Background}\label{sec:background}

\subsection{AI in Media and Public Interest}

Most of the analysis of AI Narratives focuses on the English-speaking world, while narratives---and their impacts---from the Global North might not be the same in the Global South, as ideas surrounding the term "Artificial Intelligence" vary across cultures and languages~\parencites[4]{caveHowWorldSees2023}[]{caveMeaningsAICrossCultural2023}. While some transnational similarities can be seen, which might come from having similar economic development and political systems, there is no single shared imaginary of AI. Local concerns around AI vary, and the news in different countries display both utopian and dystopian narratives~\parencites{wangAIAnxietyComparing2023}.

Narratives about AI might not accurately reflect its state. These imaginings are interpreted by researchers and regulators and can lead to wrong notions of AI~\parencites[5]{caveIntroductionImaginingAI2020}. AI Narratives have the power to shape the development of the technology~\parencites[5,7]{caveHowWorldSees2023}[]{dillonWhatAIResearchers2023}. For example, framing related to a "race" in AI development might lead to actors overlooking safety precautions~\parencites[2]{caveAIRaceStrategic2018}.

This kind of power makes it valuable for different stakeholders to try to control these narratives to suit their interests, and tech companies invest resources to foster public acceptance of AI~\parencites[6]{caveIntroductionImaginingAI2020}. These efforts are reflected in the media. An analysis of six mainstream outlets in the United Kingdom shows that industry leads the debate in the news over academic studies or political speeches. As a result, UK news often fails to acknowledge some of the concerns about AI. It is presented as a solution for societal problems (such as cancer and renewable energy), and journalists rarely question whether that is the case~\parencites[1--4]{brennenIndustryledDebateHow2018}. This representation might lead to expectations over what \textcite[24]{brennenWhatExpectWhen2022} describe as pseudo-artificial general intelligence (alluding to a hypothesized system that can mimic human intelligence), by overestimating the capacity of the technology and neglecting its costs and risks. Another analysis on the UK pointed to a tendency to use a sensationalistic tone in headlines, oscillating between optimistic solutionism and warnings of systemic dangers~\parencite{roeWhatTheyreNot2023}.

The optimistic tone is influenced by the sources quoted in the articles~\parencites[29]{brennenWhatExpectWhen2022}. By often listening to experts who support AI, for instance, Canadian media reduces the controversiality of representations of this area's future from the coverage. These experts also displayed a tendency to frame AI's controversial topics in favor of the institutions for which they spoke~\parencites[3-4]{dandurandFreezingOutLegacy2023}. This speaks to the importance of analyzing the entities represented in news texts, which will be explored in this article. A lack of critical perspective might lead to relevant issues, such as societal concerns and power dynamics around AI, being left underscrutinized in a news landscape that favors potential over current reality, as noted in Canada~\parencite{guillaumedandurandTrainingNewsCoverage2022}.

The optimistic tone has been demonstrated empirically by analyzing 30 years of \textit{The New York Times'} AI coverage (1986-2016), particularly on its prospects in healthcare and education, along with a marked increase in AI-related articles since 2009~\parencites{fastLongTermTrendsPublic2017}. This coverage shows a rising concern regarding the potential for losing control over AI systems, ethical dilemmas, and adverse employment impacts. 

\begin{wrapfigure}{r}{0.55\textwidth}
    \centering
    \includegraphics[width=0.53\textwidth]{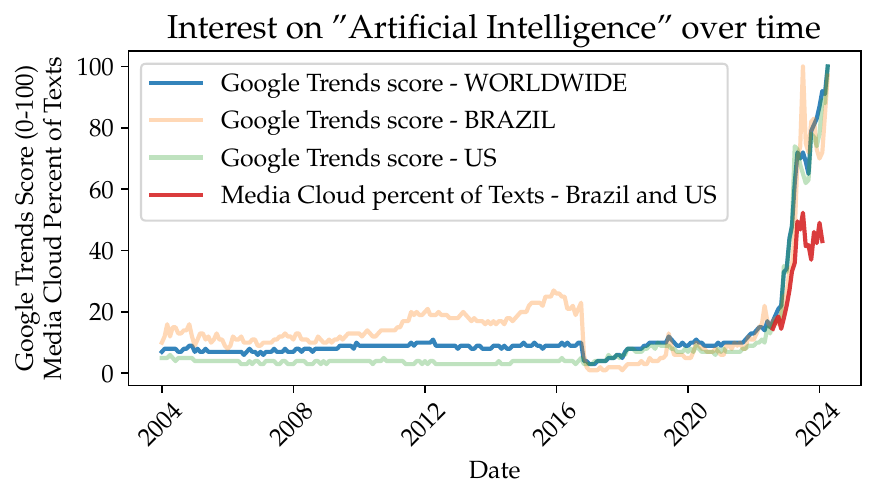}
    \caption{Google Trends scores for AI and Media Cloud ratio of news mentioning AI in the US and Brazil.}
    \label{fig:trends}
\end{wrapfigure}

Interest in AI is high in both wealthy and underprivileged countries. Google Trends\footnote{A Google service that shows user interest in particular topics in their search platform over time.} indicates that worldwide users' interest in AI is the highest since 2004, when the data starts, after taking off in 2022. The same pattern applies to each of the ten most populous countries\footnote{China, India, United States, Indonesia, Pakistan, Nigeria, Brazil, Bangladesh, Russia, and Mexico~\parencites{CensusBureauCurrent}.}~\parencites{GoogleTrends2024}, which comprise a wide range of locations and socioeconomic realities.

The interest spike can be seen in the news. Media Cloud\footnote{Service that collects news from RSS feeds worldwide~\parencites{robertsMediaCloudMassive2021}.} data, available from January 2022, shows an increase over time in the ratio of texts that mention AI in US and Brazilian media (Figure~\ref{fig:trends}). Comparing the ratio of articles about AI to the scores from Google Trends in each country and worldwide shows a high correlation in every combination (Pearson's \(r\) ranged from .908 to .939, \(p < 0.001\)). This sort of relationship is to be expected. Agenda-setting and agenda-building theories have shown that the news sets the public debate and is influenced by it when picking stories to cover~\parencites{weaverMediaAgendaWho2014,Tandoc2021,macgregorTrackingOnlineAudience2007}.

\subsection{Science and Technology Journalism}

Science journalism has been widely analyzed in literature and may provide a valuable background for understanding the conditions of producing technology-related news. It is similar to tech journalism for disseminating new, often technical, findings. Works about AI~\parencites[2,8]{brennenIndustryledDebateHow2018} and about science journalism in Brazil that refer to news outlets' "Science and Tech" sections~\parencites{massaraniUmRaioDos2013}.

Part of the increased AI itch can only be scratched by S\&T journalism, especially in the Global South, given its lack of alternative science communication platforms, such as science museums, cafes, and exhibitions~\parencites[2,12]{nguyenScienceJournalismDevelopment2019}. However, specialist reporters are expensive and, thus, rare in newsrooms as the overall number of journalists is in decline~\parencites[15--16]{trenchRoutledgeHandbookPublic2021}. Those who keep their jobs in this environment are pressured to increase their content output, diminishing quality~\parencites[50]{mikes.schaferHowChangingMedia2017}. In Brazil, the journalism industry faces financial struggles, which led to layoffs and worse working conditions, especially since the 2010s~\parencites{magalimoserUmJornalismoSem2019}. These struggles are known to other countries in the Global South and lead to generalist reporters being assigned to write about these complex subjects, resulting in less critical output~\parencites[9]{nguyenScienceJournalismDevelopment2019}. 

\section{Methods}\label{sec:methods}

\subsection{Analyzing texts with computational methods}

Computational methods provide an effective way of analyzing trends in large quantities of texts. In a news media context, these have been explored to study biases~\parencites[394--395]{hamborgAutomatedIdentificationMedia2019}[]{hamborgRevealingMediaBias2023}. By grouping articles based on their word choices (document clustering), researchers can identify the different topics or themes in the corpus, which has been used to spot the main themes in news about an issue on the news~\parencites[397]{hamborgAutomatedIdentificationMedia2019}[]{chipidzaTopicAnalysisTraditional2022}. 

Computational Grounded Theory (CGT) offers a structured methodology that integrates human analytical skills and computational efficiency, facilitating a reproducible approach to analyzing vast corpora. It employs a three-step pipeline. Initially, unsupervised machine learning techniques like topic modeling are used to identify emergent patterns within textual data. Subsequently, it engages in a deeper, interpretative examination of the data, guided by the insights from the first step. Finally, it uses additional computational techniques to validate the identified patterns~\parencites{nelsonComputationalGroundedTheory2020}. CGT has been applied, for example, to map the main topics of discussions about AI in medical communities on Reddit~\parencites{weberHowMedicalProfessionals2024}.

\subsection{Data Analysis}

\begin{wrapfigure}{r}{0.55\textwidth}
    \centering
    \includegraphics[width=0.53\textwidth]{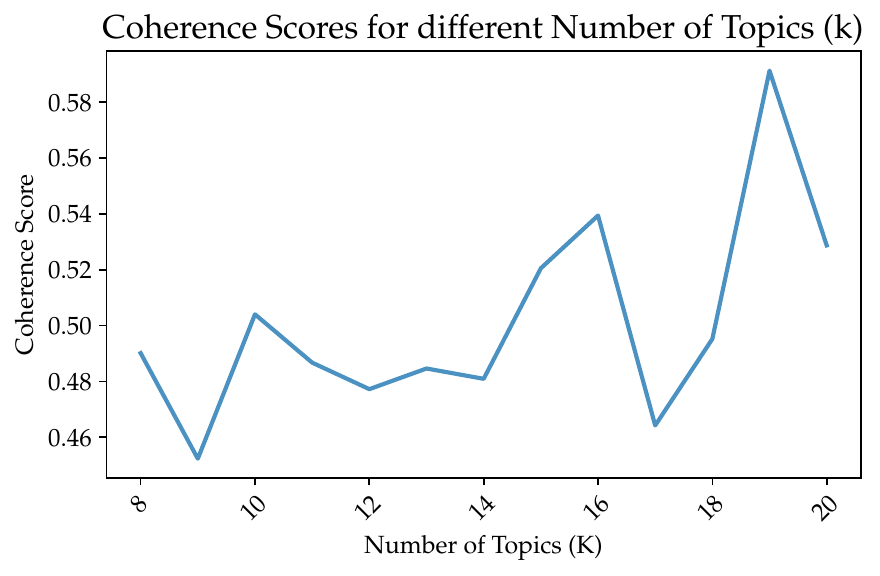}
    \caption{Number of topics (\(k\)) and coherence score (\(c_v\)). Highest: \(k=19, c_v=0.5913\).}
    \label{fig:topics}
\end{wrapfigure}

The analysis queried relevant Brazilian collections on Media Cloud\footnote{Sample query: \url{https://bit.ly/4cRuNc7}}~\parencites{robertsMediaCloudMassive2021} for texts mentioning "\textsc{Artificial Intelligence}" published between July 1st, 2023, and February 29th, 2024. They were then filtered only to contain data from a sample of 13 news outlets, curated to include popular outlets that published about AI regularly and that had different characteristics. A list of outlets and classifications is available in Appendix~\ref{sec:outlets}. This granted an initial dataset of 7,514 texts. After filtering only texts in which AI was central (Section~\ref{sec:filtering}), the analysis ran on 3,560 texts, described in Appendix~\ref{sec:corpus_stats}.

"Artificial Intelligence" was deemed a relevant keyword as it is an umbrella term for multiple applications in the realm of automating functions of the human brain~\parencites[4]{caveIntroductionImaginingAI2020}. Mentions of this term in media capture its different contexts. Similar research has also investigated broader keyword lists~\parencites[390]{recchiaFallRiseAI2020}. However, under the scope of this study, using additional terms would add complexity without bringing relevant insight. Expanding search terms in Media Cloud's set of 167 national media sources for Brazil to include "\textsc{robots}", "deep learning", and "machine learning" adds 107 texts to the 15,307 initially found.

Additional information for the relevant texts was gathered using the \texttt{news-please} Python library~\parencites{hamborgNewspleaseGenericNews2017}. A full description of the texts' preprocessing is available in Appendix~\ref{sec:preprocessing}.

\paragraph{Latent Dirichlet Allocation (LDA)} The texts were initially processed using LDA. This approach is used to discover topics in large textual datasets in journalism research~\parencites[85]{guntherWordCountsTopic2018b}[398]{hamborgAutomatedIdentificationMedia2019}. This study used LDA modeling from the \texttt{Gensim} library in Python~\parencites{radimrehurekSoftwareFrameworkTopic2010}. Optimal values for the alpha and eta hyperparameters\footnote{Alpha and eta influence the distribution of topics in documents and words in topics, respectively. A higher alpha value allows documents to be associated with more topics. A higher eta means topics are made up of more words~\parencites{LDAAlphaBeta2015}.} were found through a grid search for maximizing coherence\footnote{Coherence is an automated measure to rate the quality of topics computed by the model~\parencites{roderExploringSpaceTopic2015}.} using \( c_v \), the best-performing metric to analyze LDA coherence~\parencites[406]{roderExploringSpaceTopic2015}. Alpha was set to \textit{0.1} and eta to \textit{1}, allowing for different levels of topic prevalence across documents~\parencites{LDAAlphaBeta2015}. \( c_v \) was also used to identify optimal passes and iterations\footnote{Passes and iterations refer to the number of times the algorithm goes through the corpus and each document~\parencites{ghanoumTopicModellingPython2021}.}, set to \textit{30} and \textit{300}, respectively. 

The number of topics (\(k\)) was also defined to maximize coherence. Every number between 8 and 20 (Figure~\ref{fig:topics}) was tested with the optimal hyperparameters. Other \(k\) values were tested previously under different settings, including values between 2 and 7, 30, and 50, but those never yielded relevant results.

The highest coherence score (\(c_v = 0.5913\)) was met with \textit{19} topics, and these made sense regarding human interpretability. The modeling filtered out words that appeared too often in the corpus (over 95\% of texts) and too rarely (less than two texts). This led to the absence of the term "\textsc{artificial intelligence}" from the topics since it features in every article.

\paragraph{Deep reading} The findings from the first step were validated by a computationally guided deep reading, in which the 10 most representative texts for each identified topic were qualitatively analyzed~\parencites[25]{nelsonComputationalGroundedTheory2020}. Representativeness is based on the proportion of words in the text that are strongly associated with the topic according to the LDA model~\parencites[25--26]{nelsonComputationalGroundedTheory2020}.

\paragraph{BERTopic and Named-Entity Recognition (NER)} Results were validated with a more sophisticated unsupervised machine learning topic modeling with BERTopic~\parencites{grootendorstBERTopicNeuralTopic2022}. It was set to infer the topic division, provided that each bucket included at least 40 texts, leading to \textit{17} topics, including a special '-1' category, a catch-all for less confidently classified documents. The process was supported by a BERT model trained in Portuguese~\parencites{souzaPortugueseNamedEntity2020,fabiosouzaBERTimbauPretrainedBERT2020}.

For NER, Python's \texttt{Spacy} library was used along with \texttt{pt\_core\_news\_lg}, its best-performing pipeline for this task in Portuguese~\parencites{Portuguese2023}. This identified people, places, and institutions in the corpus.

\subsection{Filtering relevant content}\label{sec:filtering}

A text mentioning AI does not make it \textit{about} AI. Term Frequency - Inverse Document Frequency (TF-IDF) was used to spot articles in which AI was a relevant term~\parencites[516--517]{saltonTermweightingApproachesAutomatic1988}. TF-IDF was applied to generate the weight of "\textsc{Artificial Intelligence}" in each text. Then, the corpus was filtered to remove the content that fell below the median of this score. The median proved to be the best threshold after manual inspection, as it was the lowest value to include only texts in which AI was a central topic. Articles that cited AI in the title were included regardless of their TF-IDF score.

\section{Results}\label{sec:results}

\subsection{RQ1: Main topics in Brazilian news about AI}

\subsubsection{Word Count Analysis}\label{subsec:wordcount}

Word Count and N-Gram analyses show the prominence of terms related to the private sector (industry) such as "\textsc{company}," "Google," and "ChatGPT." Other common words in the AI lexicon, like "\textsc{model}", "\textsc{data}", and "\textsc{tool}", also appear. A list of the ten most frequent words is available in Appendix~\ref{sec:common_words}.

N-Gram analysis\footnote{Counting how often groups of words appear together.}, in parts, reflect the topics that drove the days with most AI-related stories published (Section~\ref{subsec:timeline}). Interestingly, however, the most common four-gram is "\textit{The New York Times}" with 119 mentions. These appear under different contexts, some as the object of stories about the newspaper's lawsuit against OpenAI~\parencites{michaelm.grynbaumNewYorkTimes2023}, others have it as the source of the stories, both from its licensing service and from texts that mention their coverage. "\textsc{United States}" is also often cited, reinforcing the international aspect of this coverage.

The ten most frequent bi-, tri-, and four-grams are available in Appendix~\ref{sec:ngrams}.

\begin{table}[htbp]
    \centering
    \begin{tabular}{@{}lccc@{}}
    \toprule
    \textbf{Category} \\ Topic description & \textbf{Topic Number} & \textbf{Top-5 Keywords} & \textbf{No. of Texts} \\ 
    \midrule
    \textbf{Culture} & & & \textbf{223} \\
    New Beatles Song & 1 & music, editor, autor, paul, beatl & 35 \\
    Jabuti & 3 & premi, livr, jabut, ilustr, frankenstein & 23 \\
    Hollywood Strike & 8 & ator, grev, estudi, sindicat, roteir & 85 \\
    Movies & 10 & film, resistenc, histor, missa, cinem & 48 \\
    Arts, Literature, Religion & 14 & livr, human, net, autor, amazon & 32 \\
    \midrule
    \textbf{Economy} & & & \textbf{1574} \\
    Finance & 11 & empres, us, merc, nvid, bilho & 342 \\
    Sam Altman & 16 & opena, altman, empres, musk, diss & 143 \\
    Uses in Work, Professional & 17 & tecnolog, empres, dad, desenvolv, human & 1089 \\
    \midrule
    \textbf{Products} & & & \textbf{818} \\
    Cell phones & 2 & galaxy, samsung, celul, linh, cam & 131 \\
    Hardware & 4 & intel, process, cor, amd, nucl & 55 \\
    New products, How-tos & 13 & googl, public, chatgpt, ferrament, continu & 632 \\
    \midrule
    \textbf{Society} & & & \textbf{927} \\
    Elections, Regulation & 7 & uso, eleitoral, conteud, eleico, president & 290 \\
    Students Deepfake & 6 & alun, escol, podcasts, colegi, adolescent & 27 \\
    Dilemmas and Misuse & 12 & vid, cri, imagens, diss, imag & 524 \\
    AI in Healthcare & 19 & medic, saud, pacient, doenc, pesquis & 86 \\
    \midrule
    \textbf{Discarded} & & & \textbf{18} \\
    Elis Regina ad & 5 & elis, rit, regin, mar, volkswagen & 4 \\
    Canaltech Podcast & 9 & canaltech, podcast, cuneiform, cathi, gustav & 12 \\
    Events Schedule & 15 & trilh, georgiev, inov, outubr, fmi & 2 \\
    Unknown & 18 & phot, orbit, wang, world, salsich & 0 \\
    \midrule
    \textbf{Overall} & & & \textbf{3560} \\ % CALCULATED SUM OF ALL TEXTS
    \bottomrule
    \end{tabular}
    \caption{LDA: Detailed Classification of Topics}
    \label{tab:topic_classification} % LABEL FOR REFERENCE
\end{table}

\subsubsection{Topic Modelling}\label{subsec:topicmodelling}

The LDA analysis divided the content into 19 topics. Four of those can be excluded as they include too few texts to be considered (\(n\le12\)). They refer to very specific cases, such as coverage of a podcast series published by \textit{Canaltech}, one of the analyzed sources. The remaining 15 can then be grouped into 4 larger categories: Culture, Economy, Products, and Society (Table~\ref{tab:topic_classification}). These were identified after an analysis of how the topics could be clustered during the deep reading stage, considering their overarching themes and news sections in which they would fit.

\begin{figure}[htbp]
    \centering
    \includegraphics[width=\textwidth]{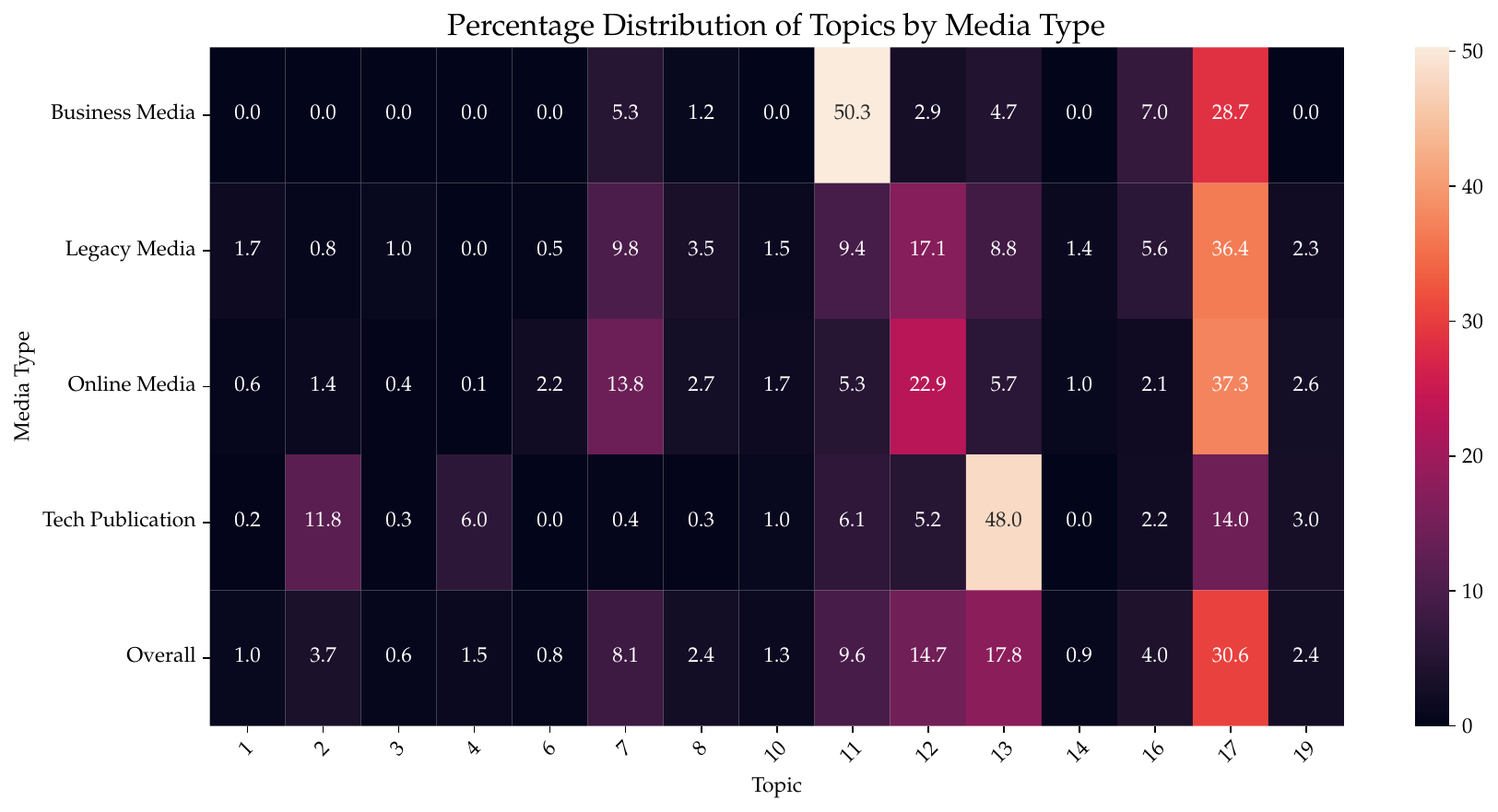}
    \caption{LDA: Distribution of articles underscores the thematic alignment of different media types with certain topics. Lighter colors correspond to a higher percentage of texts in a topic within a media category. Topics with too few articles (5, 9, 15, 18) were removed.}
    \label{fig:distribution}
\end{figure}

The topic distribution is not uniform (Figure~\ref{fig:distribution}). Topic 17, related to AI's usage at work, is the most popular overall and ranks high in every media type category. It is least present in tech publications (where it still ranks second). Tech media drives the coverage of products. Topic 7, related to the use of AI in the elections (the most prominently covered local AI issue), was only barely present in the two tech publications this analysis focused on, while that was the third most relevant issue for online and legacy media. Unsurprisingly, finance-related news makes up most of the business media's coverage.

\paragraph{Culture}

The Culture category includes texts from three topics strictly related to specific events. Topic 1 covers the new song by The Beatles, which used AI to improve old recordings. Topic 8 includes the Hollywood strikes. These had actors and writers protesting, among other issues, for protection against the use of AI in their business. Topic 3 is about a literary prize disqualifying a book illustration over it being created with the help of AI tools. 

Overall, the top-ranking texts for each topic show that this episodic coverage lacks depth in discussing AI, referring to it as a generic entity without analyzing broader issues. In the literary prize case, for example, coverage does not progress from narrating the events and some of their repercussions. It fails to discuss whether AI-generated art should be considered art or its role as a creative tool~\parencites{AIgeneratedArtActually,g.m.trujillojr.AIArtArt2022}. Topic 10 includes movies that have AI in their plots. Topic 14 is a potluck of different applications.

\paragraph{Economy}

\begin{figure}[htbp]
    \centering
    \includegraphics[width=\textwidth]{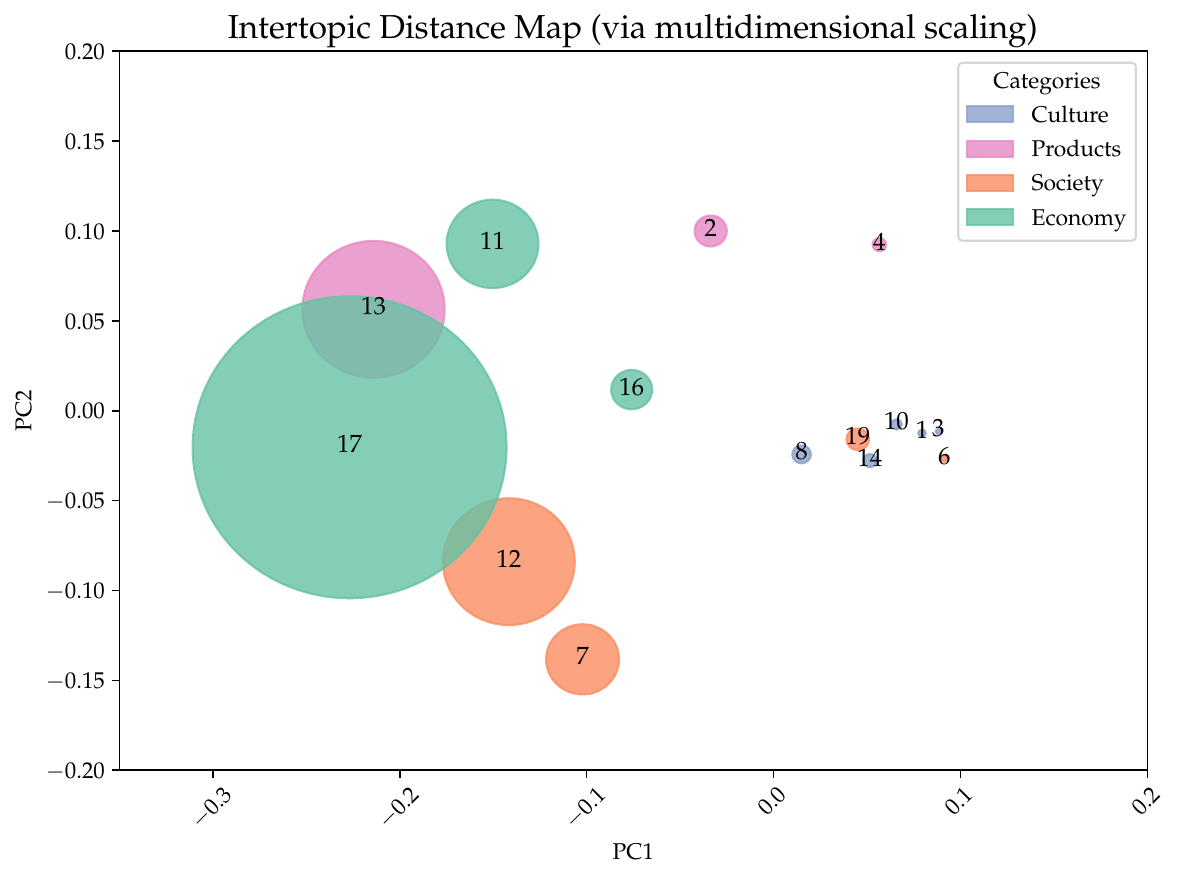}
    \caption{The Intertopic Distance Map displays the thematic spaces of LDA-generated topics in a two-dimensional plane, where each bubble's size reflects the topic's prevalence. Distances between bubbles represent differentiation between them. Closer bubbles have more thematic overlap.}
    \label{fig:intertopic}
\end{figure}

Topic 17, the largest in volume of texts, is related to work. Its articles discuss trends in the job market, highlighting a forecasted uptick in positions involving AI, which relates to another frequent subject in this topic, applying AI in the workplace and expectations for the future. The modeling suggests a proximity between topics 17 and 13 (new products and how-tos) (Figure~\ref{fig:intertopic}). This can be verified by texts explaining how to leverage AI for work or how to deal with automated selection processes, including some discussions of dealing with AI biases. There are discussions on job displacement caused by AI, though that was not a dominant subject.

Topic 16 articles are about Sam Altman's ousting and reinstatement as OpenAI's CEO in November 2023, with frequent recapitulations of events. Topic 11 focuses on the stock prices of companies related to AI, including coverage of Nvidia's market value skyrocketing. In these cases, AI is often cited as a justification for financial success. 

\paragraph{Products}

The second largest topic overall, 13, includes announcements of new products and features, explanations about how to use AI tools, and which tool to use in different situations. These explanations are connected to the launches themselves. For example, on February 13\textsuperscript{th}, 2024, \textit{Canaltech} published an explainer about Google Gemini~\parencites{haasQueIAGemini2024}, the chatbot launched a few days earlier to replace Bard~\parencites{sissiehsiaoBardBecomesGemini2024}.

Topic 4 relates to the launches of computer processors or laptops. Topic 2 mentions the launch of smartphones described by their makers as having AI functionalities (for editing pictures, for example). AI capabilities are sometimes vaguely mentioned without explaining what they are.

\paragraph{Society}

Most topics in this category show a concern around deepfakes\footnote{Realistic images generated or edited with AI.}. Topic 12 encapsulates AI's ethical dilemmas and its potential for wrongdoings when generating fake images. These include testimonies of people who saw their faces in forged videos used for scams, the usage of AI to bring back images and voices of late artists, and a TikTok trend of generating gruesome videos of kids. Most texts analyzed in this category were originally published by \textit{BBC World Service}.

The largest topic in the Society category, 7, focuses on the legal debate, mainly relating to using AI during Brazil's (then upcoming) 2024 elections. Most content is driven by the rules defined by the Superior Electoral Court (TSE), which required candidates to be transparent about their AI usage and banned deepfakes and chatbots that communicate directly with voters~\parencites{fernandavivasEleicoes2024Campanha2024}. Topic 7 also includes articles about bills in Congress to regulate AI, again focusing on deepfakes.

Topic 6 relates to school students who used AI to make pornographic deepfakes of their colleagues. Most texts refer to a case that happened in Rio. A similar episode happened in the state of Pernambuco a few days later, so the topic continued to be covered.

Topic 18 relates to AI in healthcare and adopts an optimistic tone. Its texts focus on the repercussions of research on using AI for supporting diagnostics or preventing illnesses. 

\subsubsection{Timeline}\label{subsec:timeline}

Science coverage adheres to the journalistic structure and norms, making it episodic (i.e., mostly based on events)~\parencites[19]{trenchRoutledgeHandbookPublic2021}[8]{nguyenScienceJournalismDevelopment2019}. This motivates an analysis of the output of AI content in Brazilian news over time to investigate whether specific events drove increased activity. The data indicates this is the case, supporting a pattern spotted in topic modeling (Section~\ref{subsec:topicmodelling}).

An analysis of pairwise Pearson correlation coefficients (\(r\)) on the number of texts published daily by the ten most frequent outlets in the corpus shows a positive relationship in every case with a statistically significant relationship (Pearson's \(r\) range: .374 to .727). This points to increased activity periods being shared among outlets, a pattern consistent with episodic behavior. When an event happens, various outlets react to it, leading to an uptick. A correlation matrix is available in Appendix~\ref{sec:correlation_matrix}. 

\begin{figure}[htbp]
    \centering
    \includegraphics[width=\textwidth]{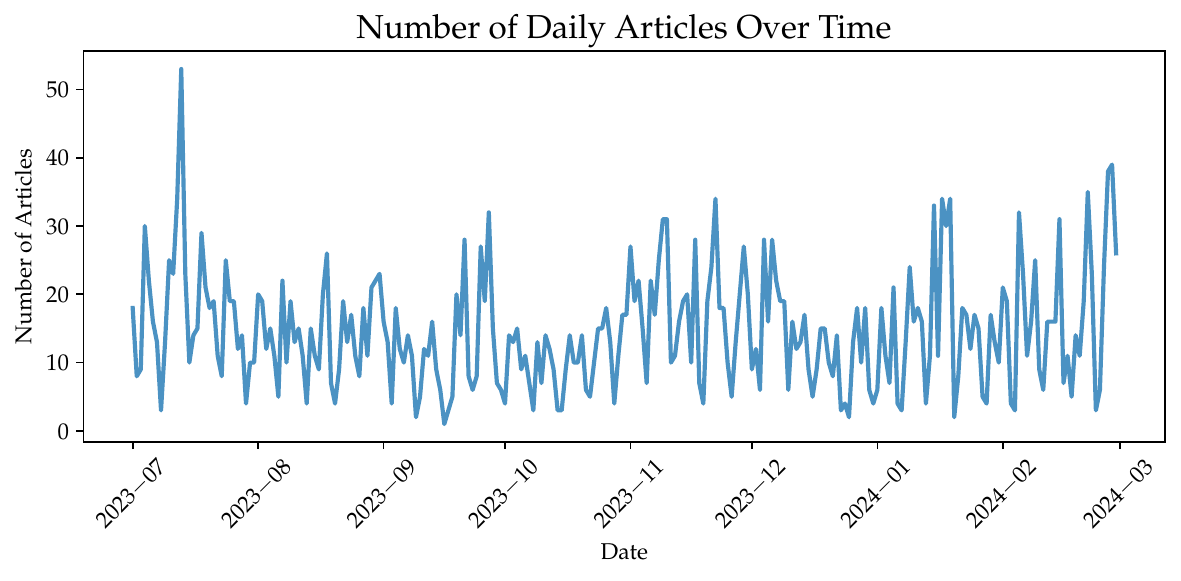}
    \caption{Daily volume of articles about AI displays surges depending on particular news events.}
    \label{fig:daily_articles}
\end{figure}

On average, 14.7 (median 14) articles were published daily (Figure~\ref{fig:daily_articles}). Four dates are outliers\footnote{$\text{Outlier if } x > Q_3 + 1.5 \times \text{IQR}$}: 13.Jul.2023, 22.Feb.2024, 27.Feb.2024, and 28.Feb.2024. The articles from each of those days were analyzed independently. In each of these news cycles, there was one clear event that drove the increase. The leading event would relate to the largest share of texts from each day (ranging from 36\% to 56\% of what was published on that date). 

On February 27\textsuperscript{th} and 28\textsuperscript{th}, 2024, the news cycle reacted to Brazil's Superior Electoral Court (TSE) creating rules for using AI in the 2024 elections. The peak activity was on July 13\textsuperscript{th}, 2023, when 19 of the 53 analyzed articles (36\%) were related to the launch of Google Bard in Brazil. On February 22\textsuperscript{nd}, 2024, Nvidia dominated the finance headlines after increasing its stock market value by US\$ 277 billion.

\begin{figure}[htbp]
    \centering
    \includegraphics[width=\textwidth]{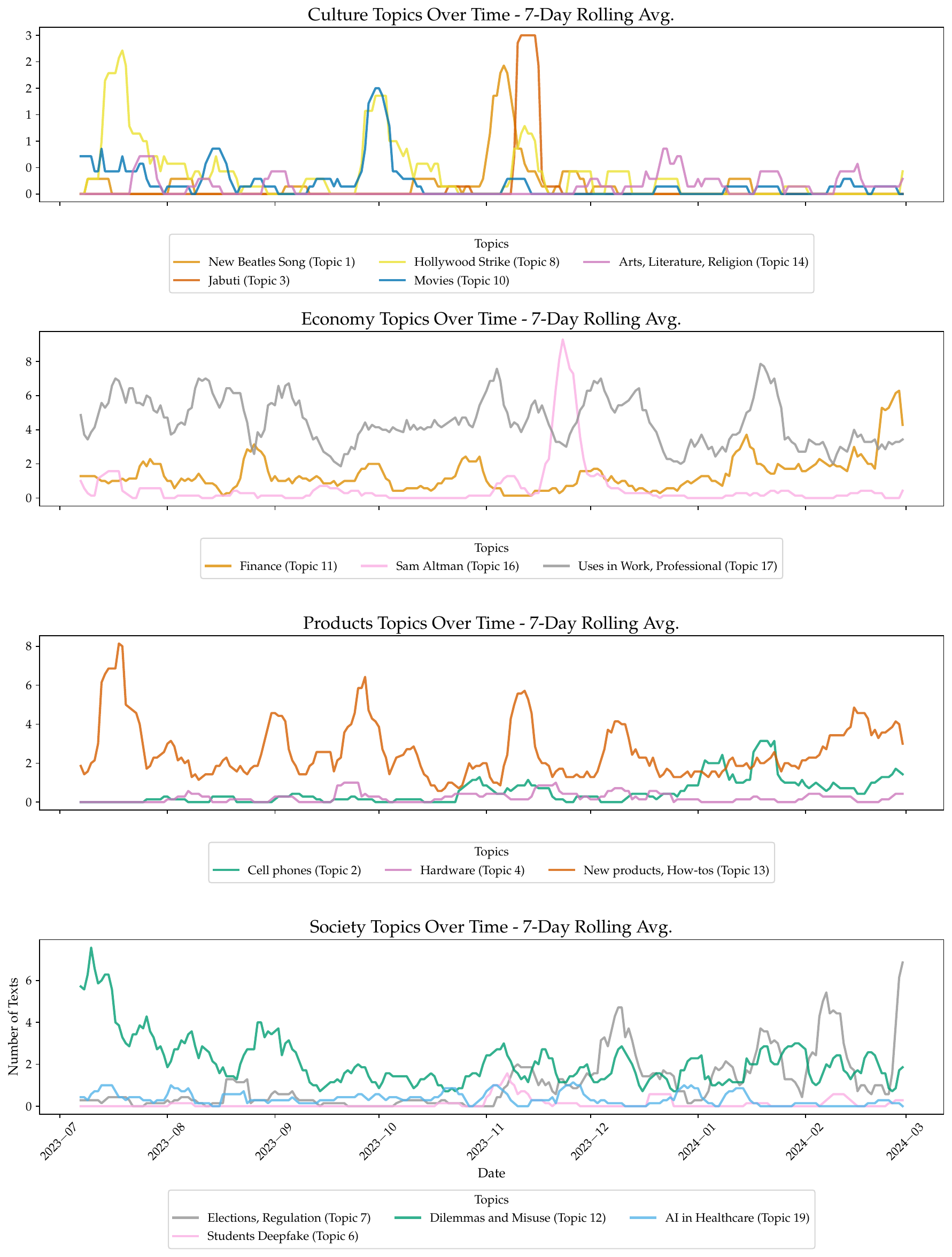}
    \caption{LDA: Number of texts published for each topic over time often show ups and downs, rather than consistent lines, reflecting episodic coverage. Topics with too few articles (5, 9, 15, 18) were removed.}
    \label{fig:topic_timeline}
\end{figure}

The episodic behavior can also be seen in the timelines of texts published per topic (Figure~\ref{fig:topic_timeline}). Most topics show an intermittent pattern even when diluting the variation over a 7-day rolling average, which accounts for natural variation in the news (publishing less on weekends, for example). Topic 12, about using AI to generate fake images, is the only societal debate on AI that could sustain some consistency.

\subsubsection{Similar topic patterns in BERTopic}\label{subsec:bert_topic_classification}

\begin{table}[htbp]
    \centering
    \begin{tabular}{@{}lccc@{}}
    \toprule
    \textbf{Category} \\ Topic description & \textbf{Topic Number} & \textbf{Top-5 Keywords} & \textbf{No. of Texts} \\ 
    \midrule
    \textbf{Culture} & & & \textbf{402} \\
    Celebrities & 4 & elis, regina, maria, cantora, rita & 208 \\
    Events Schedule & 14 & evento, festival, inovacao, programacao, g1 & 75 \\
    Hollywood strike & 15 & atores, greve, estudios, sindicato, roteiristas & 73 \\
    Movies & 16 & filme, resistencia, missao, historia, filmes & 46 \\
    \midrule
    \textbf{Economy} & & & \textbf{556} \\
    Work & 2 & \makecell{empresas, dados, tecnologia,\\brasil, profissionais} & 266 \\
    Nvidia stock price & 7 & us, nvidia, bilhoes, mercado, acoes & 178 \\
    Sam Altman & 9 & altman, openai, empresa, microsoft, sam & 112 \\
    \midrule
    \textbf{Products} & & & \textbf{647} \\
    New products, How tos & 1 & google, bard, chatgpt, microsoft, ferramenta & 441 \\
    Cell phones & 10 & galaxy, s24, samsung, linha, ultra & 109 \\
    Hardware & 12 & intel, core, processadores, amd, lake & 97 \\
    \midrule
    \textbf{Society} & & & \textbf{944} \\
    Elections & 3 & eleitoral, tse, eleicoes, tribunal, uso & 228 \\
    Deepfakes & 5 & imagens, video, conteudo, policia, vitimas & 203 \\
    Regulation, International & 6 & riscos, tecnologia, seguranca, paises, disse & 188 \\
    Healthcare & 8 & saude, pacientes, cancer, diagnostico, estudo & 145 \\
    Dilemmas and Misuse & 11 & \makecell{chatgpt, humanos, humano,\\modelos, tecnologia} & 102 \\
    Copyright lawsuits & 13 & direitos, openai, autorais, times, conteudo & 78 \\
    \bottomrule
    \end{tabular}
    \caption{BERTopic: Detailed Classification of Topics}
    \label{tab:bert_topic_classification}
\end{table}

\begin{figure}[htbp]
    \centering
    \includegraphics[width=\textwidth]{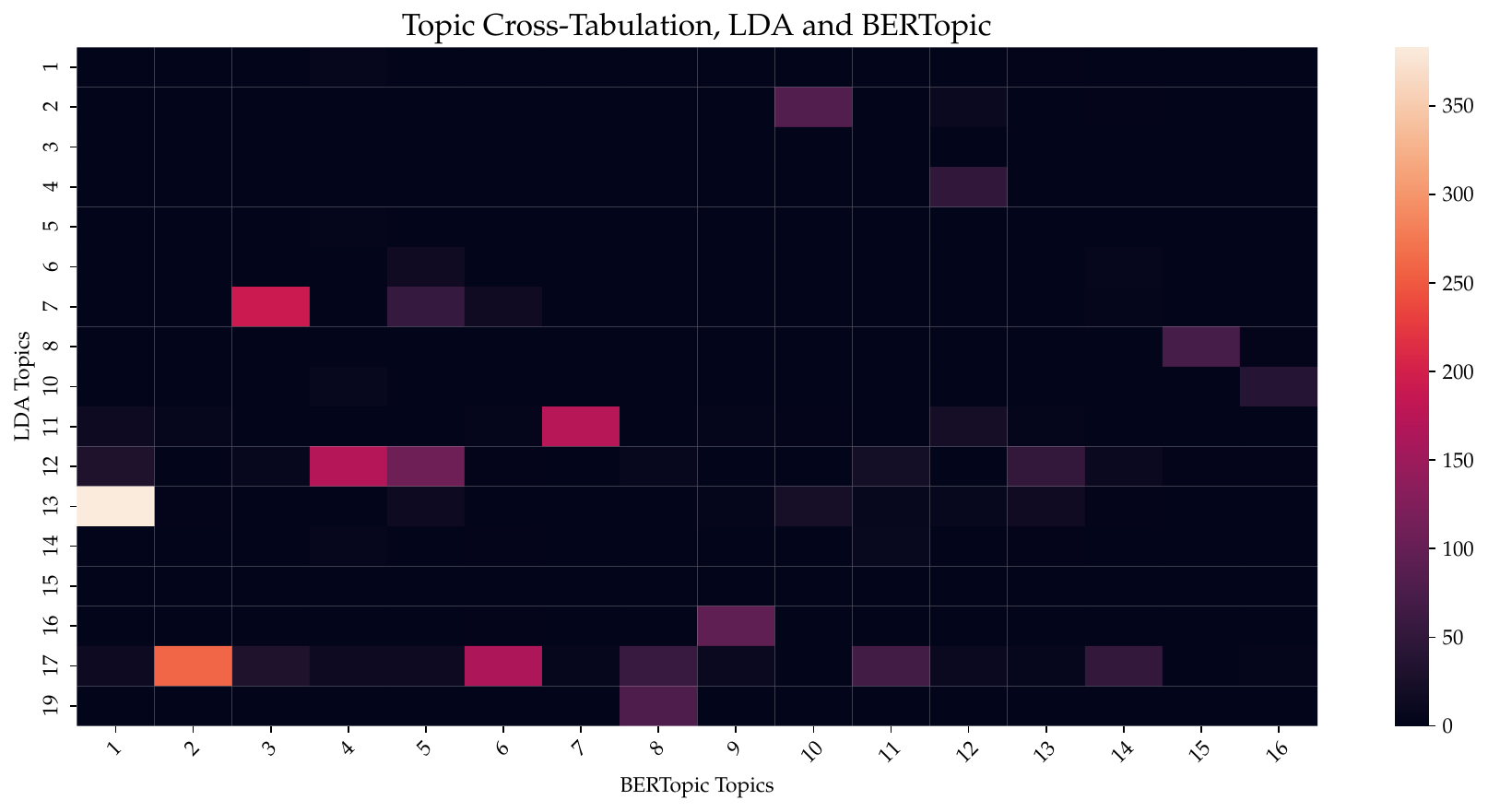}
    \caption{The heatmap shows the distribution and alignment of topics generated by LDA and BERTopic. Darker colors indicate more documents that both models have similarly categorized, suggesting agreement on the underlying thematic content. LDA topics with too few articles (5, 9, 15, 18) were removed.}
    \label{fig:crosstab_bert_lda}
\end{figure}

While the size of the topics varies, BERTopic's output can be arranged in a similar category structure to LDA's, which helps confirm some of its findings. These include topics for new products and how-tos, work-related news, elections, finance, and deepfakes (Figure~\ref{fig:crosstab_bert_lda}). An Adjusted Rand Index\footnote{Documents assigned to BERTopic's '-1' topic were excluded.} score of 0.4475 indicates a moderate agreement between the topic assignments from LDA and BERTopic.

\paragraph{Artists and International} BERTopic picks up on two additional themes in the corpus. Its Topic 4 includes news about the use of Elis Regina's image, a Brazilian singer who died in 1982, being reproduced with AI in an ad, and other news about using AI to reproduce singers' voices. News about the international AI landscape, including the US and China relations and AI regulation, appears in Topic 6.

\subsection{RQ2: Sources represented in Brazilian news about AI}\label{subsec:entities}

NER analysis confirms the prevalence of industry-related actors in the stories, as seen in topic modeling and deep reading. Fifteen of the Top 20 entities in the documents relate to big tech companies.

Looking into these entities' popularity over time helps ensure that their position in the rankings results from a pattern rather than a spike in coverage. There are moments of increased mentions for some, such as when Altman was ousted from OpenAI, but most appear widely throughout the period. 

\begin{table}[htbp]
    \centering
    \begin{tabular}{cll|cll}
    \toprule
    Rank & Entity & Count & Rank & Entity & Count \\
    \midrule
    1 & ChatGPT & 2922 & 11 & Elon Musk & 798 \\
    2 & OpenAI & 2536 & 12 & Samsung & 641 \\
    3 & Google & 2465 & 13 & China & 628 \\
    4 & Microsoft & 1944 & 14 & Amazon & 577 \\
    5 & Brasil & 1835 & 15 & TSE & 494 \\
    6 & EUA & 1744 & 16 & Internet & 471 \\
    7 & Sam Altman & 1070 & 17 & Twitter & 456 \\
    8 & Nvidia & 1033 & 18 & Instagram & 420 \\
    9 & Bard & 841 & 19 & Intel & 388 \\
    10 & Meta & 804 & 20 & Apple & 795 \\
    \bottomrule
    \end{tabular}
    \caption{Top-20 Most Popular Entities (excludes "\textsc{Artificial Intelligence}"). Some entities were grouped, such as mapping "Altman" to "Sam Altman". "EUA" means "USA".}
    \label{tab:ner-entities}
\end{table}

\begin{figure}[htbp]
    \centering
    \includegraphics[width=\textwidth]{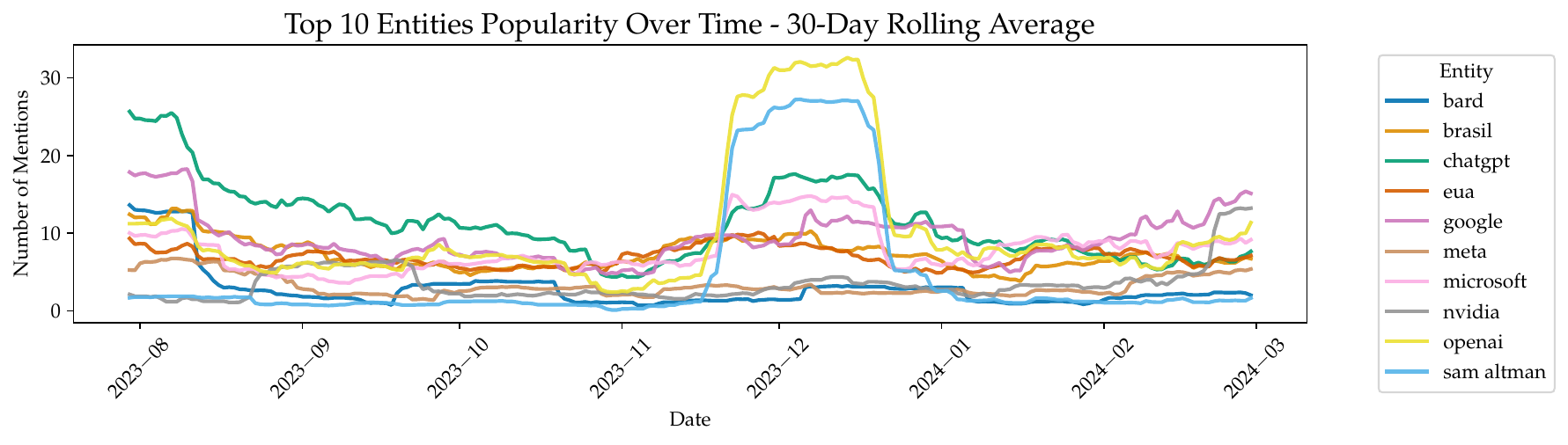}
    \caption{Mentions of the top-10 entities over time (excludes "\textsc{Artificial Intelligence}") show that most of them are regularly present in the news. "Altman" and "Sam Altman" were grouped. A rolling average (30 days) was used for easier viewing. "EUA" means "USA". }
    \label{fig:top_entities_over_time}
\end{figure}

\section{Discussion}\label{sec:discussion}

The main topics and entities mentioned in news coverage shed light on the sources of information and the actors shaping the discourse. Overall, the topic distribution reveals a focus on AI products, in the workplace and elsewhere, over deeper discussions of their implications for society, policy, or ethics. While there is coverage of topics like elections and healthcare, these are overshadowed by content related to new AI features, product launches, and corporate developments. This pattern suggests a prioritization of technological advancement and its immediate applications. This is similar to Canada, where AI is often covered in business news and with an optimistic tone~\parencites[10-11]{guillaumedandurandTrainingNewsCoverage2022}.

In Brazil, technology is connected to recent menaces to the country's democracy, especially between 2012 and 2022, with the use of disinformation online to cast doubt on the elections' integrity, promote radicalization, erode trust in media and democratic institutions, and harm political actors' reputations~\parencites{melloMaquinaOdioNotas2020,ozawaHowDisinformationWhatsApp2023}. Much of the news with a local flavor is related to deepfakes with a regulatory and elections lens, and the analysis highlights the importance of elections in the Brazilian AI debate, reflecting a unique scenario. The Superior Electoral Court (TSE) has been at the forefront of tech regulation in the country, as bills on the matter have not progressed in Congress, and relevant cases in the Supreme Court are still awaiting judgment. It is up to TSE to delve into technological progress since elections happen every two years and, thus, cannot wait for the other institutions' rulings~\parencites{moraesQueEstaPor2024}.

In some ways, Brazilian AI coverage echoes that from abroad. This is shown by \textit{The New York Times}'s popularity and by texts reproduced from the BBC dominating one of the LDA topics. While understandable, given the financial constraints of the news organizations and the fact that most AI development happens away from Brazil, heavy reliance on foreign sources in S\&T journalism leads to a disconnect with local issues~\parencites[5--7]{nguyenScienceJournalismDevelopment2019}.

Themes that are particularly relevant in the Global South are left out. Racism, which has deep roots in the country's technological history~\parencites{kingAfrofuturismoAestheticsResistance2023}, is only cited 52 times. Most refer to the electoral regulation citing racist content as an example of what should be removed by platforms~\parencites{constancarezendeTSECriaRegra2024}. Surveillance, another major concern in AI~\parencites[77]{charliebeckettReportingArtificialIntelligence2023}, is overlooked. As of April 2024, 209 projects that apply facial recognition to policing and public security have been mapped in Brazil~\parencites{MonitorReconhecimentoFacial2024}.

Like other countries in the Global South, Brazil suffers from high inequality and low employment rates~\parencites{fernandocanzianSuperricosNoBrasil2019,mohantyInequalityPerspectiveGlobal2018,zeinalatifDesempregoAltoMas2023}, making AI and automation in the workplace a major discussion point~\parencites{meiraAIIAII2023}. These are prominent topics in Brazil's AI coverage, focusing on applying AI at work rather than discussing its malaises, like job displacement. 

Digital colonialism is only briefly discussed through a lens of Brazilian companies lagging behind international competitors~\parencites{rogerioluizAmanhaPodeAcontecer2023}. There is a prevalence of US-based companies in the corpus, which provides additional evidence of these corporations' reach. In other words, this investigation points to digital colonialism being present and the media failing to pick it up. Journalism acts as a watchdog of those in power~\parencites[526--527]{norrisWatchdogJournalism2014}, and the corporations that dominate the AI world must be under this scrutiny. This idea could justify part of their mentions. However, a large share of the content reverberates their agenda, echoing the announcement of new products, for example. This relates to previous research noting that the reliance on sources related to such companies can be tied to news framings that follow their agenda~\parencites[3-4]{dandurandFreezingOutLegacy2023}.

The dominance of product coverage is even more accentuated in specialized media. The main topics in the two tech news outlets analyzed point to an avoidance of societal issues. A previous analysis of five of Brazil's most relevant specialized tech outlets found a high influx of texts, averaging 50 daily articles per website. Only \nicefrac{1}{3} of what they published, however, was related to technology. The rest included, for example, lists of products to buy and TV schedules of football matches. These are justified as efforts to stay financially afloat by generating new revenue streams and including content to capture search engine traffic~\parencites{rodrigoghedinBreveAnaliseDos2024}. This behavior, paired with the results from this paper, shows a lack of quality in Brazil's tech media when it comes to promoting deeper debates.

Time-constrained reporters rely on easier-to-digest sources, like content coming from public relations (PR) agencies, leading to stories that repeat claims from these pieces of information~\parencites[19]{brennenIndustryledDebateHow2018}[56--57]{mikes.schaferHowChangingMedia2017}. This scenario facilitates the control of the narrative by corporations, who invest a lot in public relations and are then faced with acritical coverage~\parencites[]{kapoorHowReportBetter2023}[111--112]{lakshmisivadasHowReportEffectively2021}[6]{caveIntroductionImaginingAI2020}. Furthermore, relying on the activity around AI also helps playing into the agenda of these companies, as they are the ones creating most cutting-edge systems~\parencites[46]{nestormaslejAIIndex20242024}. Critically engaging with them is a way to remediate that.

The acritical nature of this news coverage, evidenced by the topic structure and the qualitative analysis of texts on each topic, can be tied to its episodic nature and lack of experts to write about AI~\parencites[]{nguyenScienceJournalismDevelopment2019}[87]{charliebeckettReportingArtificialIntelligence2023}. Being episodic speaks to the nature of journalism and is rooted in news values that privilege recency and unexpectedness, as research has shown for over half a century~\parencites{meissner50YearsGaltung2015}. This nature, however, comes with problems, particularly when discussing complex topics like AI. The data shows significant surges in articles about specific events without necessarily sustaining an ongoing dialogue or leveraging the attention to a topic to foster debate.

\section{Conclusion}\label{sec:conclusion}

This was the first study investigating how AI is covered in Brazilian news and highlights a media landscape that is largely reactive, commercially influenced, and episodic. By analyzing Brazilian media, this addresses the gap in research on AI coverage outside the anglophone world~\parencites[5]{caveHowWorldSees2023}. This helps understand the concerns of hundreds of millions of people in a less privileged part of the globe that might not benefit from technological advancements as much as the wealthy bits~\parencites{anapaulaloboSilvioMeiraEntendimento2023,danniyuAIDivideGlobal2023}.

The main topics in the news about AI in Brazil show a predominance of coverage related to applying AI in work and discussing new products. Moreover, the most cited people and organizations show a high concentration of sources related to the industry, echoing a pattern seen abroad~\parencites{brennenIndustryledDebateHow2018,guillaumedandurandTrainingNewsCoverage2022}. This reinforces the agendas of powerful corporate stakeholders, thereby narrowing the public's understanding of AI's broader societal impacts. The significant focus on electoral integrity and deepfakes in Brazilian media reflects the country's recent history with disinformation and its impact on democracy~\parencites{melloMaquinaOdioNotas2020,ozawaHowDisinformationWhatsApp2023}. These findings underscore that while there are common themes in AI coverage globally, local contexts heavily influence the specifics of media narratives.

These findings highlight the need for more critical and diverse perspectives in AI journalism. This could involve training journalists to critically engage with AI technologies and their societal impacts, as a solid foundation is needed for this kind of expert coverage~\parencites{lakshmisivadasHowReportEffectively2021}. Also, since some of the problems arise from episodic coverage, investing in issue-based and analytical reporting, which does not rely on news events to bring up relevant discussions regularly, can be an alternative to foster more profound debate.

\subsection*{Disclosure statement}

The authors report no competing interests. 

\section*{Author contributions}

RH designed the study, RH performed the analyses, and both authors assisted in the revision of the manuscript and the refinement of its arguments.

\printbibliography

\newpage
\setcounter{page}{1}
\pagenumbering{Roman}
\begin{appendices}
\section*{Appendix}
\section{List of analyzed media outlets and their classification}\label{sec:outlets}

\begin{table}[htbp]
    \centering
    \begin{tabular}{lllll}
    \hline
    \textbf{URL} & \textbf{Outlet} & \textbf{Suboutlet} & \textbf{Classification} & \textbf{Audience} \\
    \hline
    canaltech.com.br & Canaltech & Canaltech & Tech Publication & National \\
    g1.globo.com & Globo & G1 & Online Media & National \\
    www1.folha.uol.com.br & Folha & Folha de S.Paulo & Legacy Media & National \\
    oglobo.globo.com & Globo & O Globo & Legacy Media & National \\
    istoe.com.br & Istoe & Istoe & Legacy Media & National \\
    otempo.com.br & O Tempo & O Tempo & Legacy Media & State \\
    noticias.uol.com.br & UOL & UOL Notícias & Online Media & National \\
    infomoney.com.br & InfoMoney & InfoMoney & Business Media & National \\
    metropoles.com & Metrópoles & Metrópoles & Online Media & National \\
    tecnoblog.net & Tecnoblog & Tecnoblog & Tech Publication & National \\
    correio24horas.com.br & Correio & Correio 24 Horas & Legacy Media & State \\
    uol.com.br & UOL & UOL & Online Media & National \\
    cartacapital.com.br & CartaCapital & CartaCapital & Legacy Media & National \\
    gazetadopovo.com.br & Gazeta do Povo & Gazeta do Povo & Legacy Media & State \\
    economia.uol.com.br & UOL & UOL Economia & Online Media & National \\
    cultura.uol.com.br & UOL & UOL Cultura & Online Media & National \\
    piaui.folha.uol.com.br & Piauí & Piauí & Legacy Media & National \\
    f5.folha.uol.com.br & Folha & F5 & Legacy Media & National \\
    guia.folha.uol.com.br & Folha & Guia Folha & Legacy Media & National \\
    tab.uol.com.br & UOL & UOL Tab & Online Media & National \\
    ruf.folha.uol.com.br & Folha & RUF & Legacy Media & National \\
    reporterbrasil.org.br & Repórter Brasil & Repórter Brasil & Online Media & National \\
    tilt.uol.com.br & UOL & UOL Tilt & Online Media & National \\
    educacao.uol.com.br & UOL & UOL Educação & Online Media & National \\
    \hline
    \end{tabular}
    \caption{Taxonomy of Brazilian Media Outlets}
    \label{tab:outlets}
\end{table}

\newpage
\section{List of translated Portuguese terms}\label{sec:terms}
\begin{table}[htbp]
    \centering
    \begin{tabular}{ll}
    \hline
    \textbf{English Term} & \textbf{Portuguese Term} \\
    \hline
    Artificial Intelligence & Inteligência Artificial \\
    Robots & Robôs \\
    Company & Empresa \\
    Model & Modelo \\
    Data & Dado, Dados \\
    Tool & Ferramenta \\
    To create & Criar \\
    Said & Disse \\
    United States & Estados Unidos \\
    Generative Artificial Intelligence & Inteligência Artificial Generativa \\
    \hline
    \end{tabular}
    \caption{Table of Translated Terms}
    \label{tab:terms}
\end{table}

\newpage
\section{Corpus statististics}\label{sec:corpus_stats}

\begin{table}[htbp]
    \centering
    \begin{tabular}{lr}
    \hline
    \textbf{Statistic}                & \textbf{Value} \\ \hline
    Total Documents                      & 3,560 \\
    Total Word Count                     & 1,032,468 \\
    Average Words per Document           & 290.02 \\
    Median Words per Document            & 235.0 \\
    Mode Words per Document              & 221 \\
    Vocabulary Size                      & 33,356 \\
    \hline
    \end{tabular}
    \caption{Basic Statistics of the Text Corpus - After Stemming and Removing Stopwords}
    \label{tab:corpus_stats_stemmed}
\end{table}

\begin{table}[htbp]
    \centering
    \begin{tabular}{lr}
    \hline
    \textbf{Statistic}                & \textbf{Value} \\ \hline
    Total Documents                      & 3,560 \\
    Total Word Count                     & 1,032,468 \\
    Average Words per Document           & 290.02 \\
    Median Words per Document            & 235.0 \\
    Mode Words per Document              & 221 \\
    Vocabulary Size                      & 57,391 \\
    \hline
    \end{tabular}
    \caption{Basic Statistics of the Text Corpus - After Removing Stopwords, Before Stemming}
    \label{tab:corpus_stats_stopwords}
\end{table}

\newpage
\section{Preprocessing}\label{sec:preprocessing}

Data from Media Cloud, including thousands of sources, was initially filtered to keep only the list chosen for this analysis. The process combined looking into those in the top half in terms of their Open Pagerank\footnote{An open-source score that reflects a website's popularity.\parencites{WhatOpenPageRank2023}} scores, to then pick the most relevant ones. Since Media Cloud data came from three collections, they had overlapping sources. Thus, duplicated news articles were found and were removed. Next, outlets were classified according to Table~\ref{tab:outlets}.

Texts were then filtered by their length. No outliers at the higher end of the length were found. The shortest 1\% of texts were removed as manual inspection showed that these were not news articles but rather short texts referring readers to some other content (introducing a podcast episode, for example). During this process, text length was analyzed per outlet to assert that there were no systemic errors leading to incomplete data collection.

Abbreviations of "\textsc{Artificial Intelligence}" ("\textit{IA}" in Portuguese) were replaced with the full term. Words were set to lowercase. Special characters, punctuation, and stopwords were removed. This process led to repeated instances of the term "\textsc{artificial intelligence} due to writing that originally said "\textsc{Artificial Intelligence (AI)}". These were replaced to keep only one occurrence of the term.

Later, a stemmed version of the texts was also generated. The analysis often used both versions, stemmed and not stemmed, to pick the one that was easier to understand.

N-gram analysis was used to spot repeated segments cluttering the text, such as systematically included excerpts asking readers to subscribe to a newsletter or to follow the news outlet on social media. The top 30 bi-gram, tri-gram, and four-grams were analyzed iteratively with the removal of such texts until no such occurrences appeared.

The texts were inspected to clear any potential leftover HTML or Javascript segments in the content. None was found. The Python package \texttt{langdetect} was run on a sample of each text to assert all of them were in Portuguese.

Finally, \textit{simhashing}\footnote{Technique that converts texts to a hash (alphanumeric representation of the content) and then uses it to compare them for similarity. Similar texts will have similar hashes. This technique requires less computation than pairwise comparing each full text.\parencites{ottenSimHashUltimateGuide2023}} was used to detect other duplicated texts (perhaps with a minimal difference that would prevent them from being found in the first deduplication). This led to the removal of 25 texts from the corpus.

\newpage
\section{Most Common Words}\label{sec:common_words}

\begin{table}[ht]
    \centering
    \begin{tabular}{lc}
    \hline
    \textbf{Word} & \textbf{Frequency} \\
  	\hline
  	inteligenc & 22,040 \\
	artificial & 21,708 \\
	empres & 7,744 \\
	tecnolog & 7,304 \\
	diss & 4,387 \\
	cri & 3,798 \\
	dad & 3,673 \\
	ferrament & 3,553 \\
	uso & 3,425 \\
	model & 3,276 \\
	\hline
    \end{tabular}
    \caption{Most Common Words in the Corpus After Stemming and Removing Stopwords.}
    \label{tab:common_words_stopwords_stemmed}
\end{table}

\begin{table}[ht]
    \centering
    \begin{tabular}{lc}
    \hline
    \textbf{Word} & \textbf{Frequency} \\
  	\hline
  	inteligencia & 21,875 \\
	artificial & 21,664 \\
	tecnologia & 5,381 \\
	empresa & 4,485 \\
	uso & 3,425 \\
	dados & 3,369 \\
	empresas & 3,244 \\
	google & 3,127 \\
	chatgpt & 3,054 \\
	disse & 3,014 \\
	\hline
    \end{tabular}
    \caption{Most Common Words in the Corpus After Removing Stopwords, Before Stemming.}
    \label{tab:common_words_stopwords_stopwords}
\end{table}

\newpage
\section{Most Common N-Grams}\label{sec:ngrams}

\begin{table}[ht]
    \centering
    \begin{tabular}{lr}
    \hline
    \textbf{N-gram} & \textbf{Frequency} \\ \hline	inteligenc artificial & 21,520 \\
  artificial gener & 1,598 \\
  uso inteligenc & 1,143 \\
  red soc & 956 \\
  estad unid & 870 \\
  ferrament inteligenc & 769 \\
  model inteligenc & 569 \\
  galaxy s24 & 549 \\
  model linguag & 517 \\
  ger inteligenc & 512 \\
  \hline
    \end{tabular}
    \caption{Top 10 Bi-grams After Stemming and Removing Stopwords}
    \label{tab:top_2_grams_stemmed_text}
\end{table}

\begin{table}[ht]
    \centering
    \begin{tabular}{lr}
    \hline
    \textbf{N-gram} & \textbf{Frequency} \\ \hline	inteligenc artificial gener & 1,596 \\
  uso inteligenc artificial & 1,133 \\
  ferrament inteligenc artificial & 768 \\
  model inteligenc artificial & 566 \\
  ger inteligenc artificial & 508 \\
  tecnolog inteligenc artificial & 403 \\
  inteligenc artificial cri & 292 \\
  desenvolv inteligenc artificial & 279 \\
  recurs inteligenc artificial & 273 \\
  sistem inteligenc artificial & 265 \\
  \hline
    \end{tabular}
    \caption{Top 10 Tri-grams After Stemming and Removing Stopwords}
    \label{tab:top_3_grams_stemmed_text}
\end{table}

\begin{table}[ht]
    \centering
    \begin{tabular}{lr}
    \hline
    \textbf{N-gram} & \textbf{Frequency} \\ \hline	the new york tim & 118 \\
  conteud ger inteligenc artificial & 111 \\
  tribunal superior eleitoral tse & 96 \\
  ferrament inteligenc artificial gener & 85 \\
  model inteligenc artificial gener & 64 \\
  trein model inteligenc artificial & 64 \\
  imagens ger inteligenc artificial & 55 \\
  suprem tribunal federal stf & 50 \\
  inteligenc artificial gener chatgpt & 48 \\
  luiz inaci lul silv & 46 \\
  \hline
    \end{tabular}
    \caption{Top 10 Four-grams After Stemming and Removing Stopwords}
    \label{tab:top_4_grams_stemmed_text}
\end{table}

\begin{table}[ht]
    \centering
    \begin{tabular}{lr}
    \hline
    \textbf{N-gram} & \textbf{Frequency} \\ \hline	inteligencia artificial & 21,519 \\
  artificial generativa & 1,570 \\
  uso inteligencia & 1,136 \\
  redes sociais & 951 \\
  estados unidos & 870 \\
  galaxy s24 & 549 \\
  ferramentas inteligencia & 514 \\
  sam altman & 383 \\
  direitos autorais & 358 \\
  modelos inteligencia & 355 \\
  \hline
    \end{tabular}
    \caption{Top 10 Bi-grams After Removing Stopwords, Before Stemming}
    \label{tab:top_2_grams_stopwords_text}
\end{table}

\begin{table}[ht]
    \centering
    \begin{tabular}{lr}
    \hline
    \textbf{N-gram} & \textbf{Frequency} \\ \hline	inteligencia artificial generativa & 1,568 \\
  uso inteligencia artificial & 1,133 \\
  ferramentas inteligencia artificial & 514 \\
  modelos inteligencia artificial & 354 \\
  sistemas inteligencia artificial & 264 \\
  ferramenta inteligencia artificial & 253 \\
  recursos inteligencia artificial & 247 \\
  new york times & 242 \\
  tecnologia inteligencia artificial & 218 \\
  modelo inteligencia artificial & 211 \\
  \hline
    \end{tabular}
    \caption{Top 10 Tri-grams After Removing Stopwords, Before Stemming}
    \label{tab:top_3_grams_stopwords_text}
\end{table}

\begin{table}[ht]
    \centering
    \begin{tabular}{lr}
    \hline
    \textbf{N-gram} & \textbf{Frequency} \\ \hline	the new york times & 118 \\
  tribunal superior eleitoral tse & 96 \\
  conteudo gerado inteligencia artificial & 72 \\
  ferramentas inteligencia artificial generativa & 58 \\
  imagens geradas inteligencia artificial & 55 \\
  treinar modelos inteligencia artificial & 51 \\
  supremo tribunal federal stf & 50 \\
  luiz inacio lula silva & 46 \\
  inteligencia artificial generativa chatgpt & 44 \\
  uso inteligencia artificial eleicoes & 44 \\
  \hline
    \end{tabular}
    \caption{Top 10 Four-grams After Removing Stopwords, Before Stemming}
    \label{tab:top_4_grams_stopwords_text}
\end{table}

\newpage
\begin{landscape}
\section{Correlation matrix}\label{sec:correlation_matrix}

\begin{table}[htbp]
    \centering
\begin{tabular}{lllllllllll}
\toprule
 & Canaltech & Correio & Folha & Globo & InfoMoney & Istoe & Metrópoles & O Tempo & Tecnoblog & UOL \\
 &  &  &  &  &  &  &  &  &  &  \\
\midrule
Canaltech & 1.000 & 0.294 & 0.049 & 0.200 & 0.308 & 0.310 & 0.188 & 0.284 & 0.424* & 0.174 \\
Correio & 0.294 & 1.000 & 0.292 & -0.169 & 0.151 & 0.168 & 0.287 & 0.198 & 0.459** & 0.615*** \\
Folha & 0.049 & 0.292 & 1.000 & 0.056 & 0.574*** & 0.252 & -0.150 & 0.312 & 0.374* & 0.562*** \\
Globo & 0.200 & -0.169 & 0.056 & 1.000 & -0.118 & 0.727*** & -0.020 & 0.527*** & -0.153 & -0.190 \\
InfoMoney & 0.308 & 0.151 & 0.574*** & -0.118 & 1.000 & 0.138 & -0.130 & -0.043 & 0.457** & 0.464** \\
Istoe & 0.310 & 0.168 & 0.252 & 0.727*** & 0.138 & 1.000 & -0.052 & 0.474** & 0.289 & 0.286 \\
Metrópoles & 0.188 & 0.287 & -0.150 & -0.020 & -0.130 & -0.052 & 1.000 & 0.322 & 0.183 & 0.035 \\
O Tempo & 0.284 & 0.198 & 0.312 & 0.527*** & -0.043 & 0.474** & 0.322 & 1.000 & 0.228 & 0.237 \\
Tecnoblog & 0.424* & 0.459** & 0.374* & -0.153 & 0.457** & 0.289 & 0.183 & 0.228 & 1.000 & 0.530*** \\
UOL & 0.174 & 0.615*** & 0.562*** & -0.190 & 0.464** & 0.286 & 0.035 & 0.237 & 0.530*** & 1.000 \\
\bottomrule
\end{tabular}
\caption{Pairwise Pearson correlation coefficients (\(r\)) between Top-10 most frequent outlets in the corpus with significance levels (\(*= p<0.05, **=p<0.01, ***=p<0.001.\))}
\label{tab:correlation_matrix}
\end{table}
\end{landscape}

\end{appendices}

\end{document}